\documentclass[conference]{IEEEtran}
\IEEEoverridecommandlockouts
\usepackage{cite}
\usepackage{amsmath,amssymb,amsfonts}
\usepackage{algorithmic}
\usepackage{graphicx}
\usepackage{textcomp}
\usepackage{xcolor}
\def\BibTeX{{\rm B\kern-.05em{\sc i\kern-.025em b}\kern-.08em
    T\kern-.1667em\lower.7ex\hbox{E}\kern-.125emX}}
\begin{document}

\title{Intrusion Detection System in Smart Home Network Using Bidirectional LSTM and Convolutional Neural Networks Hybrid Model\\
}

\author{\IEEEauthorblockN{Nelly Elsayed, Zaghloul Saad Zaghloul, Sylvia Worlali Azumah, Chengcheng Li}
\IEEEauthorblockA{\textit{School of Information Technology} \\
\textit{University of Cincinnati}\\
Cincinnati, Ohio, United Stated\\
nelly.elsayed@uc.edu, elsayezs@ucmail.uc.edu, azumahsw@mail.uc.edu,li2cc@ucmail.uc.edu}
}

\maketitle

\begin{abstract}
Internet of Things (IoT) allowed smart homes to
improve the quality and the comfort of our daily lives. However, these conveniences introduced several security concerns that increase rapidly. IoT devices, smart home hubs, and gateway raise various security risks.
The smart home gateways act as a centralized point of communication between
the IoT devices, which can create a backdoor into network data for
hackers. One of the common and effective ways to detect such attacks is intrusion detection in the network traffic. In this paper, we proposed an intrusion detection system (IDS) to detect anomalies in a smart home network using a bidirectional long short-term memory (BiLSTM) and convolutional neural network (CNN) hybrid model. The BiLSTM recurrent behavior provides the intrusion detection model to preserve the learned information through time, and the CNN extracts perfectly the data features. The proposed model can be applied to any smart home network gateway. 
\end{abstract}


\section{Introduction}
IoT devices become essential to various users as it provides users with devices control, data receiving and sharing through the internet without 
Using IoT devices nowadays has become very helpful to various users' human intervention~\cite{wortmann2015internet,atzori2010internet}. The smart home is an example of building an automation system for different home devices that can communicate to provide comfort, convenience, support, and security for the home users~\cite{harper2006inside,brdiczka2008learning}. Smart home designed to provide a unique ecosystem  that called Web of Things that provides the interconnection between different types of embedded devices with tags to integrate them into a Web application using the Web standards~\cite{zeng2011web,guinard2011internet}. Popularly known IoT devices used in smart homes include Alex, Google Home, and video doorbells. Others include biometric cybersecurity scanners, fitness trackers, apple watches, medical sensors, and many more~\cite{lopatovska2019talk,nijholt2008google,mashal2019makes,anomalyLSTM}.

IoT gateway is hardware physical or a virtual device that can receive data from IoT sensors to be sent to the fog or the cloud~\cite{chen2011brief, iorga2018fog, hayes2008cloud}.
IoT gateway is hardware physical or a virtual device that can receive data from IoT sensors to be sent to the fog or the cloud~\cite{chen2011brief, iorga2018fog, hayes2008cloud}. The IoT allows local processing and storage and autonomously controls field devices based on data inputs by sensors~\cite{kang2018experimental}.

\begin{figure}[t]
	\centering
	\includegraphics[width=8.75cm, height = 3.5cm]{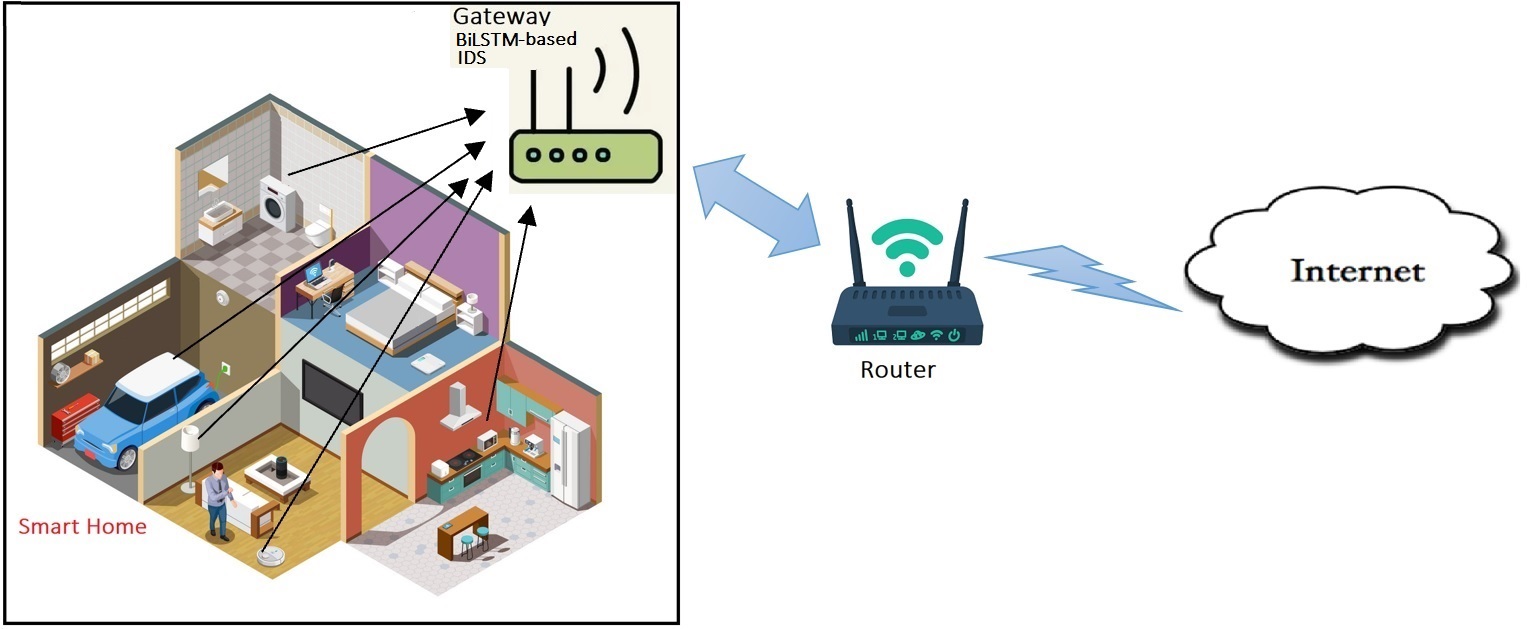}
	\caption{The smart home using the proposed BiLSTM-CNN hybrid IDS model for anomaly intrusion detection.}
	\label{IDS-System}
\end{figure}

The particular contribution of this paper is a novel solution model based on the bidirectional long short-term memory (BiLSTM) and convolutional neural network (CNN) to detecting intrusions in the smart home by leverages the IoT Gateway to detect the occurrence of IoT network anomalies in a smart home.  The proposed model will monitor, detect, and trigger actions based on the anomalies from network traffic into and within the IoT network.
The proposed BiLSTM-CNN hybrid intrusion detection system (IDS) placement is shown in Figure~\ref{IDS-System} where the BiLSTM+CNN based model is employed in the intermediate stage between the smart home gateway and the Internet.

\section{Background}




\begin{figure}
    \centering
	\includegraphics[width=7cm, height=4.5cm]{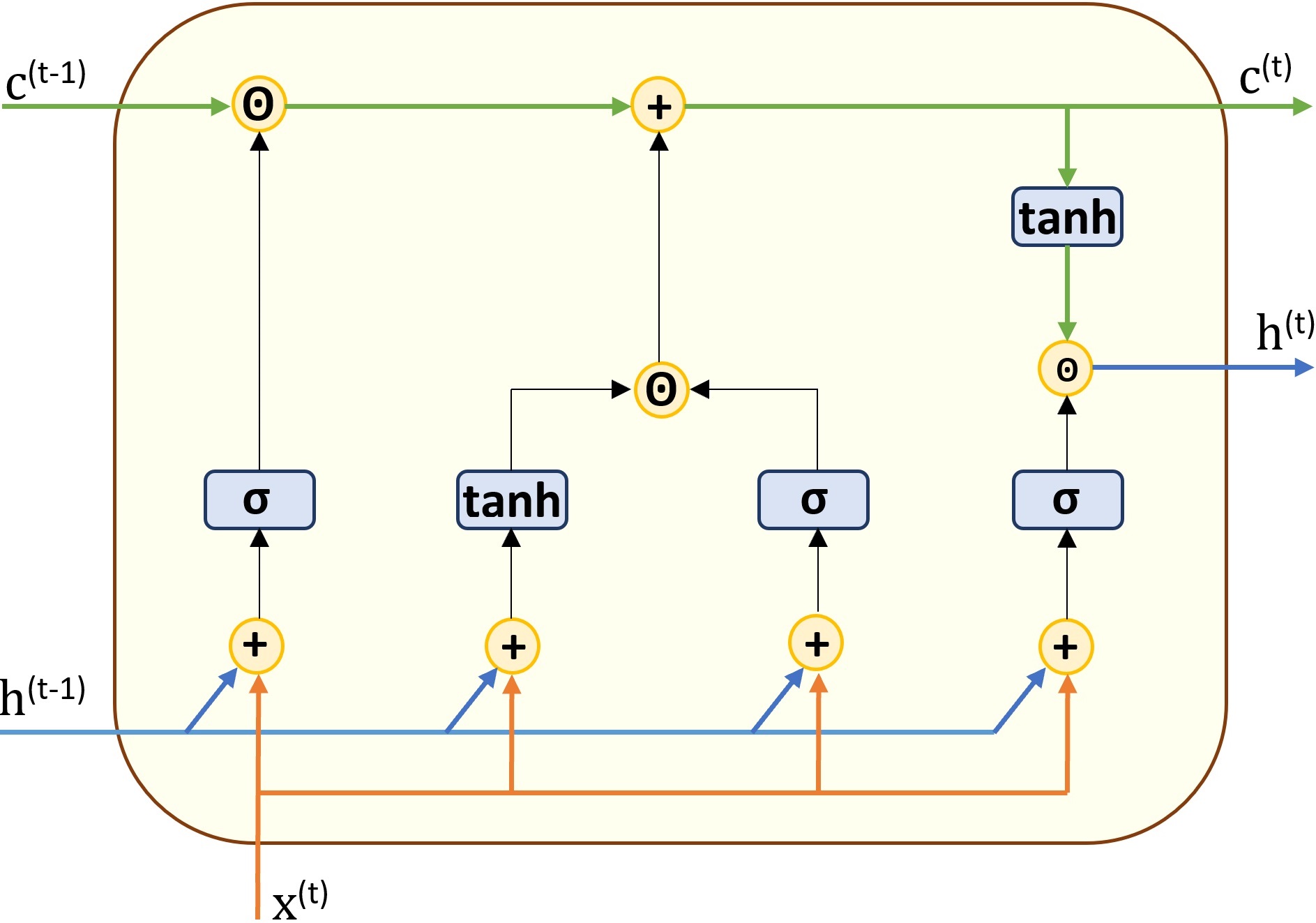}
	\caption{The LSTM architecture block~\cite{elsayed2019gated}.}
	\label{LSTM_block}
\end{figure} 

\begin{figure}
    \centering
	\includegraphics[width= 7.75cm, height=4.5cm]{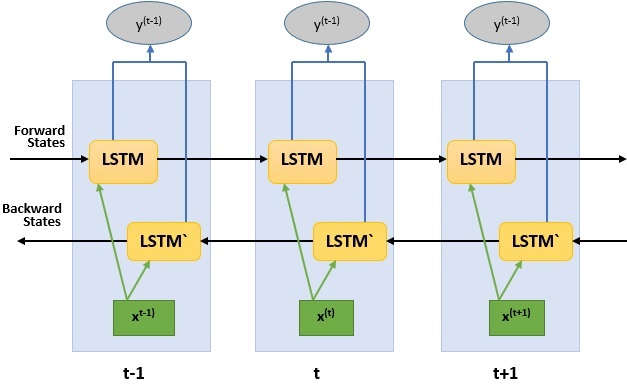}
	\caption{The general architecture of the bidirectional LSTM (BiLSTM) shown in three time steps unfolding.}
	\label{biLSTM}
\end{figure} 

\subsection{Bidirectional LSTM}

The bidirectional recurrent neural network (BRNN) was first developed by Schuster and Paliwal~\cite{schuster1997bidirectional} in 1997, where two hidden layers of the recurrent architecture in the opposite direction are connected to produce an output. This bidirectional behavior increases the input data flexibility for the recurrent architecture. Moreover, the recurrent bidirectional network increases the reachability of future state inputs to the current state and does not require the input data to be fixed prior to the training process~\cite{salehinejad2017recent}.  

In this paper, we used the long short-term memory (LSTM) as the recurrent unit of the bidirectional recurrent architecture as it overcomes the vanishing/exploding gradient problem that occures in the recurrent neural network (RNN). In addition, Graves et al.~\cite{graves2005framewise} using the LSTM in the bidirectional architecture showed a significance improvment in the classification accuracy. The LSTM architecture is shown if Figure~\ref{LSTM_block} where $c^{(t)}$, $h^{(t)}$, and $x^{(t)}$ are the memory state cell, LSTM output at time $t$, and the input at time $t$, respectively. The symbol $\odot$ denotes the element-wise (Hadamard) multiplication. The $\mathit{tanh}$ is the hyperbolic tangent function~\cite{medsker2001recurrent} and $\sigma$ is the logestic sigmoid function~\cite{schtickzelle1981pierre}. The LSTM components values are calculated as follows:
\begin{align}
i^{(t)}&= \sigma(W_{xi} x^{(t)} +U_{hi}  h^{(t-1)}+ b_i)\label{eqn:i_lstm}\\ 
g^{(t)}&= \mathrm{tanh}(W_{xg} x^{(t)} +U_{hg}  h^{(t-1)}+ b_g)\label{eqn:g_lstm}\\ 
f^{(t)}&=\sigma(W_{xf} x^{(t)} +U_{hf}  h^{(t-1)} + b_f)\label{eqn:f_lstm}\\ 
o^{(t)}&=\sigma(W_{xo} x^{(t)} +U_{ho}  h^{(t-1)} + b_o)\label{eqn:o_lstm}\\ 
c^{(t)}&= f^{(t)}\odot c^{(t-1)} + i^{(t)} \odot g^{(t)}\label{eqn:s_lstm}\\
h^{(t)}&= \mathrm{tanh}(c^{(t)})\odot q^{(t)}\label{eqn:h_lstm}
\end{align}
\noindent
where $i^{(t)}$, $f^{(t)}$, and $o^{(t)}$ are the input, forget, and output gates, respectively. $g^{(t)}$, is the input-update value. $b_i$, $b_g$, $b_f$, and $b_o$ are the biases of each gate. $W$'s are the feedforward weights and and $U$'s are the recurrent weights. The model has two activation units: input-update and output activation where $\mathrm{tanh}$ activation function is the preferable function to be used~\cite{elsayed2018a}.

The general architecture of the BiLSTM in three time steps unfolding is shown at Figure~\ref{biLSTM}. The BiLSTM architecture training process is shown in Figure~\ref{biLSTM} where the Bi-LSTM computes two sequences: the forward hidden sequence \textbf{$\overrightarrow{h}$} and the backward hidden sequence \textbf{$\overleftarrow{h}$} to produce the output sequence \textbf{$y$} by iterating the forward layer ascending from time $t=1$ to $t=T$ and the hidden backward layer descending from time $t=T$ to $t=1$~\cite{graves2013hybrid}. The forward, backward and output sequences are calculated by:
\begin{align}
\overrightarrow{h}^{(t)}&= \mathcal{H}(W_{x\overrightarrow{h}} x^{(t)} +W_{\overrightarrow{h}\overrightarrow{h}}  \overrightarrow{h}^{(t-1)}+ b_{\overrightarrow{h}})\label{eqn:forward}\\ 
\overleftarrow{h}^{(t)}&= \mathcal{H}(W_{x\overleftarrow{h}} x^{(t)} +W_{\overleftarrow{h}\overleftarrow{h}}  \overleftarrow{h}^{(t+1)}+ b_{\overleftarrow{h}})\label{eqn:backward}\\ 
y^{(t)}&= W_{\overrightarrow{h}y} \overrightarrow{h}^{(t)} +W_{\overleftarrow{h}y}\overleftarrow{h}^{(t)}+ b_y)\label{eqn:output_y}
\end{align}

The bidirectional structure incorporates the temporal dynamic of the recurrent system as the model trained in both the feedforward and backward directions~\cite{schuster1997bidirectional}.

\subsection{Convolutional Neural Networks}
The Convolutional Neural Network (CNN), was first introduced by LeCun et al.~\cite{lecun1989backpropagation} in
1989 to utilizes weight sharing over grid-structured datasets
such as time series and images. The convolutional layers learn by extracting the complex feature representations from raw or little preprocessed data. The Neural convolutional networks showed a significant increase in performance improvement in various applications.


\section{Proposed BiLSTM-CNN Hybrid Model}

\begin{figure*}
	\centering
	\includegraphics[width=16cm, height = 5cm]{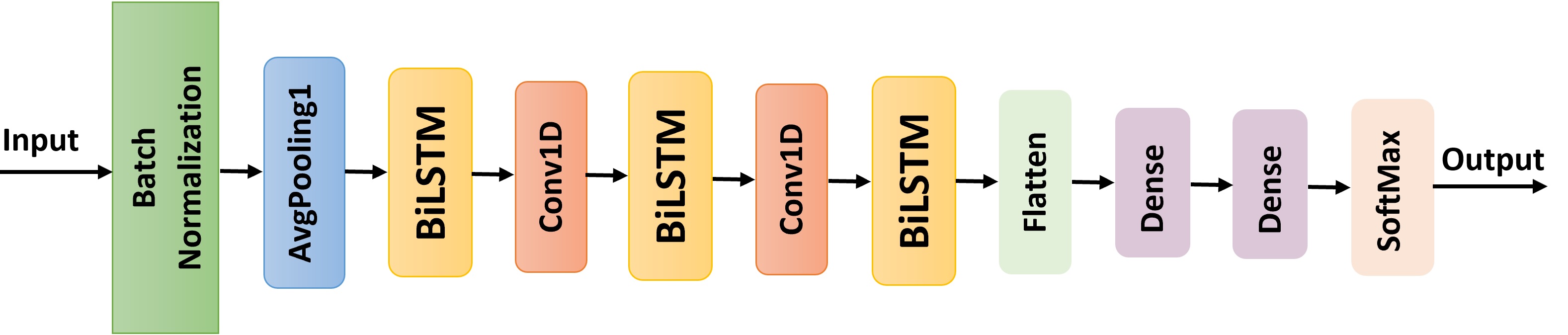}
	\caption{The proposed BiLSTM-CNN hybrid IDS model architecture.}
	\label{proposedModel}
\end{figure*} 

The proposed BiLSTM-based IDS model is shown in Figure~\ref{proposedModel}. The proposed model consists of eleven layers. The first layer is the batch normalization responsible for normalizing each input batch that fits into the model, keeping into account that the data has not been scaled or preprocessed prior to the fit into the model. The second layer is the 1D average pooling layer to reduces the overall computational cost, training parameters and prevent the model of training overfitting problem~\cite{boureau2010theoretical,elsayed2018deep}. The next layers are the BiLSTM components followed by convolutional 1D layers. These layers are responsible for learning the temporal relation between the network flow and adjust the temporal dynamics due to the BiLSTM recurrent bidirectional architecture. Moreover, these layers are the core layers for feature extraction from the network flow. Then, the Flatten layer is used to adjust the input dimensionality to the following dense layers. Finally, the SoftMax layer responsible for determining the input class, whether a normal network flow or an anomaly, requires to trigger an action toward suppressing the network traffic flow attack. Our experiments determined the crucial data features that mainly affect the anomaly-type detection to extract for the model design. These features including the originating IP address, destination IP address of connected devices, destination port number, time of captured packets in the IoT device, normal and anomaly captured packets, and the attack category. The attack categories are including Mirai, DoS, MITM ARP, normal, and scan.

\section{Empirical Results and Analysis}

\subsection{Model Implementation}
The proposed model was implemented on an Intel(R) Core(TM) i7-9750H CPU, 2.59 GHz processor, 32 GB memory, 64-bit Windows 10 OS. We used Python 3.7.3, Tensorflow 1.15.0, and Keras 2.1.0.

The proposed BiLSTM parameters where adjusted for each layer. The average pooling 1D layer used the stride of size two and pooling of size three. Each BiLSTM was unfolded into ten unfolds. The Truncated Normal function with mean $\mu = 0$ and standard deviation $\sigma = 0.05$ has been used as the recurrent weights initialization function. The 1D convolutional layers are set to 128 kernels of size three, and the he-uniform function is used for the kernel initialization. We trained the model using 100 epochs, and we set the batch size to 32.

\subsection{Dataset}
The dataset that has been used to train and test the proposed BiLSTM model is the IoT Intrusion Dataset which available on the IEEEDataPort~\cite{https://doi.org/10.21227/p5qa-bk07}. The dataset consists of 42 raw network packet files (pcap) that are captured at different time points~\cite{https://doi.org/10.21227/p5qa-bk07}. The IoT devices, namely SKT NUGU (NU 100) and EZVIZ Wi-Fi camera (C2C Mini O Plus 1080P) were used in generating IoT devices traffic~\cite{https://doi.org/10.21227/p5qa-bk07}. These devices were set up to have other peripherals connected on the same networks as the IoT devices prior to generating the dataset. The data packets were captured using a wireless network adapter that had headers. These headers were cleaned using Aircrack-ng~\cite{https://doi.org/10.21227/p5qa-bk07}. The network sniffer that has been used to capture all the attacks that were launched is the NMAP.

\begin{figure}
	\centering
	\includegraphics[width=7cm, height =3.5cm]{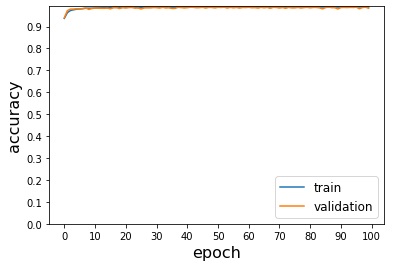}
	\caption{The proposed BiLSTM-based IDS model training vs. validation accuracy.}
	\label{accuracy}
\end{figure} 

\begin{figure}
	\centering
	\includegraphics[width=7cm, height = 3.5cm]{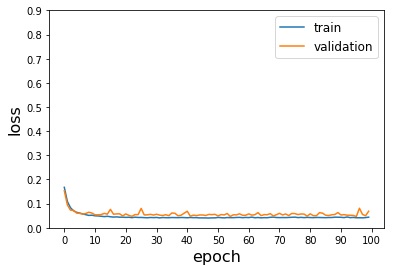}
	\caption{The proposed BiLSTM-based IDS model training vs. validation loss.}
	\label{loss}
\end{figure} 

\subsection{Results and Analysis}
The IoT Intrusion Dataset has been divided into training, validation, and testing data with the ratios 60\%, 20\%, and 20\%, respectively. Figure~\ref{accuracy} and Figure~\ref{loss} show the training versus validation accuracy and loss, respectively.
The proposed BiLSTM model performance for detection of attacks on IoT devices is measured based on four standard metrics: accuracy, recall, precision, and F1-score that can be calculated by:
\begin{align}
Accuracy &=\frac{TP + TN}{TP + TN +FP + FN}\\ 
Recall &=\frac{TP }{TP + FN}\\
Precision &=\frac{TP}{TP + FP}\\
F1 &= 2 \times \frac{Precision * Recall}{Precision + Recall}
\end{align}
\noindent
where TP ,TN, FN, and FP represents true positives, true negatives, false negatives, and false positives, respectively.

Table~\ref{table:1} shows the results of sample size $n=3$ testing trials of the proposed BiLSTM. 


\begin{table}[ht!]
\centering
\caption{The proposed BiLSTM empirical results} 
\begin{tabular}{lc} 
 \hline
 \textbf{Metrics}&\textbf{Testing Result}\\ 
 \hline
  Accuracy & 98.93\%\\
  Precision & 98.20\%\\
 Recall & 99.61\%\\ 
 F1-Score& 98.90\%\\
 \# train parameters&42,180\\
 \# all parameters&42,182\\
 \hline
\end{tabular}
\label{table:1}
\end{table}

\begin{table}[ht!]
\centering
\caption{The proposed BiLSTM empirical results} 
\begin{tabular}{lll} 
 \hline
 \textbf{Model}&\textbf{Methodology}&\textbf{Accuracy}\\ 
 \hline
  Pahl et al.~\cite{pahl2018all} & K-Means& 96.3\%\\
  Diro et al.~\cite{diro2018distributed}& ANN& 98.27\%\\
  Azumah et al.~\cite{anomalyLSTM}& LSTM &97.94\%\\
  Alrashdi et al.~\cite{alrashdi2019ad}&RF-ET&98.01\%\\
  McDermott et al.~\cite{mcdermott2018botnet}&BiLSTM-RNN&98.48\%\\
  \textbf{our model} & BiLSTM-CNN &\textbf{98.93\%}\\
 \hline
\end{tabular}
\label{table2}
\end{table}

We compared our proposed BiLSTM model to the machine learning-based state-of-the-art models' performances to detect anomalies in the IoT network traffic flow. Table~\ref{table2} shows the anomaly detection accuracy and the model primary anomaly detection methodology used. Table~\ref{table2} shows that our proposed BiLSTM-CNN based Model exceeds the state-of-the-art anomaly detection models.

\section{Conclusion}
The proposed BiLSTM-CNN hybrid model outperforms the state-of-the-art anomaly detection models for IoT devices. Moreover, it can be implemented and applied to any smart home network gateway. Furthermore, it can be connected to a decision-making alarm system that either automatically controls the smart home network or sends a notification to the homeowners/authorized members to identify any abnormalities in their smart home networks and control the situation by taking the appropriate action to mitigate the existing threat to protect their homes and data.

\section*{Acknowledgment}
This material is based upon work supported by the National Science Foundation under Grant No. (CNS-1801593). Any opinions, findings, and conclusions or recommendations expressed in this material are those of the author(s) and do not necessarily reflect the views of the National Science Foundation.

\bibliography{references_file}

\bibliographystyle{ieeetr}

\end{document}